\documentclass[]{article}
\usepackage[letterpaper]{geometry}
\usepackage{mtsummit2021}
\usepackage{times}
\usepackage{url}
\usepackage{latexsym}
\usepackage{natbib}
\usepackage{layout}
\usepackage{graphicx}
\usepackage[normalem]{ulem}
\useunder{\uline}{\ul}{}

\usepackage{multicol}

\usepackage{appendix}
\usepackage{float}
\restylefloat{table}
\usepackage{amsmath}

\begin{document}

\title{\bf Transformers for Low-Resource Languages:\\Is Féidir Linn!}

\author{\name{\bf Séamus Lankford} \hfill  \addr{seamus.lankford@adaptcentre.ie}\\
        \addr{ADAPT Centre, Department of Computing, Dublin City University, Dublin, Ireland.}
        \AND
       \name{\bf Haithem Afli} \hfill \addr{haithem.afli@adaptcentre.ie}\\
        \addr{ADAPT Centre, Department of Computer Science, Munster Technological University, Ireland.}
        \AND        
        \name{\bf Andy Way} \hfill \addr{andy.way@adaptcentre.ie}\\
        \addr{ADAPT Centre, School of Computing, Dublin City University, Dublin, Ireland.}
}
\maketitle
\pagestyle{empty}

\begin{abstract}
The Transformer model is the state-of-the-art in Machine Translation. However, in general, neural translation models often under perform on language pairs with insufficient training data. As a consequence, relatively few experiments have been carried out using this architecture on low-resource language pairs. In this study, hyperparameter optimization of Transformer models in translating the low-resource English-Irish language pair is evaluated. We demonstrate that choosing appropriate parameters leads to considerable performance improvements. Most importantly, the correct choice of subword model is shown to be the biggest driver of translation performance. SentencePiece models using both unigram and BPE approaches were appraised. Variations on model architectures included modifying the number of layers,  testing various regularisation techniques and evaluating the optimal number of heads for attention. A generic 55k DGT corpus and an in-domain 88k public admin corpus were used for evaluation. A Transformer optimized model demonstrated a BLEU score improvement of 7.8 points when compared with a baseline RNN model.  Improvements were observed across a range of metrics, including TER, indicating a substantially reduced post editing effort for Transformer optimized models with 16k BPE subword models. Bench-marked against Google Translate, our translation engines demonstrated significant improvements. The question of whether or not Transformers can be used effectively in a low-resource setting of English-Irish translation has been addressed. Is féidir linn - yes we can.
\end{abstract}

\section{Introduction}

The advent of Neural Machine Translation (NMT) has heralded an era of high-quality translations. However, these improvements have not been manifested in the translation of all languages. Large datasets are a prerequisite for high quality NMT. This works well in the context of well-resourced languages where there is an abundance of data. In the context of low-resource languages which suffer from a sparsity of data, alternative approaches must be adopted. 

An important part of this research involves developing applications and models to address the challenges of low-resource language technology. Such technology incorporates methods to address the data scarcity affecting deep learning for digital engagement of low-resource languages.  

It has been shown that an out-of-the-box NMT system, trained on English-Irish data, achieves a lower translation quality compared with using a tailored SMT system (Dowling et al, 2018). It is in this context that further research is required in the development of NMT for low-resource languages and the Irish language in particular.

Most research on choosing subword models has focused on high resource languages~\citep{ding2019call, gowda2020finding}. In the context of developing models for English to Irish translation, there are no clear recommendations on the choice of subword model types. One of the objectives in this study is to identify which type of subword model performs best in this low resource scenario.

\section{Background}
   
Native speakers of low-resource languages are often excluded from useful content since, more often than not, online content is not available to them in their language of choice. Such a digital divide and the resulting social exclusion experienced by second language speakers, such as refugees living in developed countries, has been well documented in the research literature ~\citep{macfarlane2008responses, alam2015digital}.

Research on Machine Translation (MT) in low-resource scenarios directly addresses this challenge of exclusion via pivot languages ~\citep{liu2018pivot}, and indirectly, via domain adaptation of models ~\citep{ghifary2016deep}. Breakthrough performance improvements in the area of MT have been achieved through research efforts focusing on NMT~\citep{bahdanau2014neural, cho2014properties}. Consequently, state-of-the-art (SOA) performance has been attained on multiple language pairs~\citep{bojar-etal-2017-findings, bojar-etal-2018-findings}.

\subsection{Irish Language}
The Irish language is a primary example of such a low-resource language that will benefit from this research. NMT involving Transformer model development will improve the performance in specific domains of low-resource languages. Such research will address the end of the Irish language derogation in the European Commission in 2021 \footnote{amtaweb.org/wp-content/uploads/2020/11/MT-in-EU-Overview-with-Voiceover-Andy-Way-KEYNOTE-K1.pdf}~\citep{way_2020} helping to deliver parity in support for Irish in online digital engagement.

\subsection{Hyperparameter Optimization}

Hyperparameters are employed in order to customize machine learning models such as translation models. It has been shown that machine learning performance may be improved through hyperparameter optimization (HPO) rather than just using default settings~\citep{sanders2017informing}.

The principle methods of HPO are Grid Search~\citep{montgomery2017design} and Random Search~\citep{bergstra2012random}]. Grid search is an exhaustive technique which evaluates all parameter permutations. However, as the number of features grows, the amount of data permutations grows exponentially making optimization expensive in the context of developing long running translation models. 
 
An effective, and less computationally intensive, alternative is to use random search which samples random configurations. 

\subsubsection{Recurrent Neural Networks}

Recurrent neural networks are often used for the tasks of natural language processing, speech recognition and MT. RNN models enable previous outputs to be used as inputs while having hidden states. In the context of MT, such neural networks were ideal due to their ability to process inputs of any length. Furthermore, the model sizes do not necessarily increase with the size of its input. Commonly used variants of RNN include Bidirectional (BRNN) and Deep (DRNN) architectures. However, the problem of vanishing gradients coupled with the development of attention-based algorithms often leads to Transformer models performing better than RNNs.

\subsubsection{Transformer}
The greatest improvements have been demonstrated when either the RNN or the CNN architecture is abandoned completely and replaced with an attention mechanism creating a much simpler and faster architecture known as Transformer ~\citep{vaswani2017attention}. 
Transformer models use attention to focus on previously generated tokens. The approach allows models to develop a long memory which is particularly useful in the  domain of language translation. Performance improvements to both RNN and CNN approaches may be achieved through the introduction of such attention layers in the translation architecture.

Experiments in MT tasks show such models are better in quality due to greater parallelization while requiring significantly less time to train. 

\subsection{Subword Models}

Translation, by its nature, requires an open vocabulary and the use of subword models aims to address the fixed vocabulary problem associated with NMT. Rare and unknown words are encoded as sequences of subword units. By adapting the original Byte Pair Encoding (BPE) algorithm~\citep{gage1994new}, the use of BPE submodels can improve translation performance~\citep{sennrich2015neural,kudo2018subword}. 

Designed for NMT, SentencePiece, is a language-independent subword tokenizer that provides an open-source C++ and a Python implementation for subword units. An attractive feature of the tokenizer is that SentencePiece trains subword models directly from raw sentences~\citep{kudo2018SentencePiece}.

\subsubsection{Byte Pair Encoding compared with Unigram}
BPE and unigram language models are similar in that both encode text using fewer bits but each uses a different data compression principle (dictionary vs. entropy). In principle, we would expect the same benefits with the unigram language model as with BPE. However, unigram models are often more flexible since they are probabilistic models that output multiple segmentations with their probabilities.

\begin{figure}
    \centering
    \includegraphics[width=12cm]{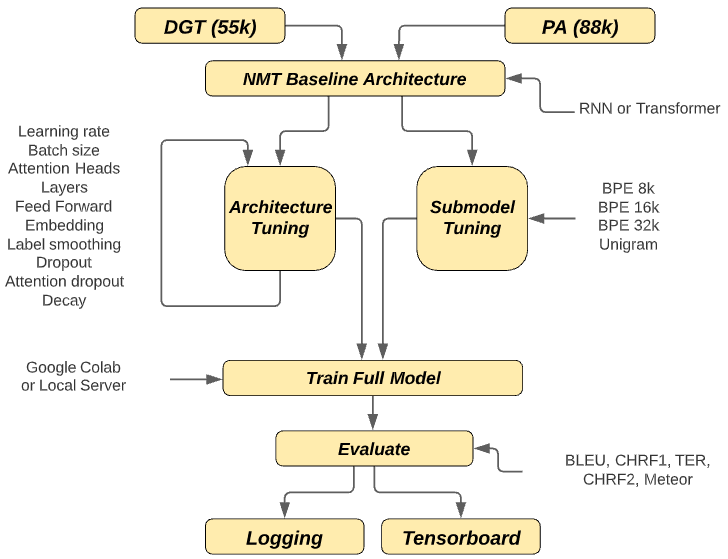}
    \caption{Proposed Approach}
    \label{fig:approach}
\end{figure}

\begin{center}
\begin{table}
\center
\begin{tabular}{ll}
\hline
\textbf{Hyperparameter} & \textbf{Values}                \\ \hline
Learning rate            & 0.1, 0.01, 0.001, \textbf{2}            \\ \hline
Batch size               & 1024, \textbf{2048},  4096, 8192       \\ \hline
Attention heads          & \textbf{2}, 4, \textbf{8}                     \\ \hline
Number of layers         & 5, \textbf{6}                           \\ \hline
Feed-forward dimension   & \textbf{2048}                           \\ \hline
Embedding dimension      & 128, \textbf{256}, 512                  \\ \hline
Label smoothing          & \textbf{0.1}, 0.3                       \\ \hline
Dropout                  & 0.1, \textbf{0.3}                       \\ \hline
Attention dropout        & \textbf{0.1}                            \\ \hline
Average Decay            & 0, \textbf{0.0001}                      \\ \hline
\end{tabular}
\caption{Hyperparameter Optimization for Transformer models. Optimal parameters are highlighted in bold. The highest performing model trained on the 55k DGT corpus uses 2 attention heads whereas the best model trained with the larger 88k PA dataset uses 8 attention heads.}
\label{tab:hpo-table}
\end{table}
\end{center}

\section{Proposed Approach}

HPO of RNN models in low-resource settings has previously demonstrated considerable performance improvements. The extent to which such optimization techniques may be applied to Transformer models in similar low-resource scenarios is evaluated as part of this study. Evaluations included modifying the number of attention heads, the number of layers and experimenting with regularization techniques such as dropout and label smoothing. Most importantly, the choice of subword model type and the vocabulary size are evaluated. 

In order to test the effectiveness of our approaches, optimization was  carried out on two English-Irish parallel datasets: a general corpus of 52k lines from the Directorate General for Translation (DGT) and an in-domain corpus of 88k lines of Public Administration (PA) data. With DGT, the test set used 1.3k lines and the development set comprised of 2.6k lines. In the case of the PA dataset, there were 1.5k lines of test data and 3k lines of validation.  All experiments involved concatenating source and target corpora to create a shared vocabulary and a shared SentencePiece subword model.  The impact of using separate source and target subword models was not explored.

The approach adopted is illustrated in Figure \ref{fig:approach}. Two baseline architectures, RNN and Transformer, are evaluated. On evaluating the hyperparameter choices for Transformer models, the values outlined in Table 1 were tested using a random search approach. A range of values for each parameter was tested using short cycles of 5k training steps. Once an optimal value, within the sampled range was identified, it was locked in for tests on subsequent parameters.

\subsection{Architecture Tuning}

Given the long training times associated with NMT, it is difficult and costly to tune systems using a conventional Grid Search approach. Therefore a Random Search approach was adopted in the HPO of our transformer models. 

With low-resource datasets, the use of smaller and fewer layers has previously been shown to improve performance ~\citep{araabi2020optimizing}. Performance of low-resource NMT has also been demonstrated to improve in cases where shallow Transformer models are adopted ~\citep{van2020optimal}. Guided by these findings, configurations were tested which varied the number of neurons in each layer and modified the number of layers used in the Transformer architecture.

The impact of regularization, by applying varying degrees of dropout to Transformer models, was evaluated. Configurations using smaller (0.1) and larger values (0.3) were applied to the output of each feed forward layer.

\subsection{Subword Models}

It has become standard practise to incorporate word segmentation approaches, such as Byte-Pair-Encoding (BPE), when developing NMT models. Previous work shows that subword models may be particularly beneficial for low-resource languages since rare words are often a problem. Reducing the number of BPE merge operations resulted in substantial improvements of 5 BLEU points (Sennrich and Zhang 2019) when tested on RNN models.

In the context of English to Irish translation, there is no clear agreement as to what constituted the best approach. Consequently, as part of this study, subword regularization techniques, involving BPE and unigram models were evaluated to determining the optimal parameters for maximising translation performance. BPE models with varying vocabulary sizes of 4k, 8k, 16k and 32k were tested.

\section{Empirical Evaluation}

\subsection{Experimental Setup}
\subsubsection{Datasets}
The performance of the Transformer and RNN approaches is evaluated on English to Irish parallel datasets. Two datasets were used in the evaluation of our models namely the publicly available DGT dataset which may be broadly categorised as generic and an in-domain dataset which focuses on public administration data.

The DGT, and its Joint Research Centre, has made available all Translation Memory (TM; i.e. sentences and their professionally produced translations) which cover all official European Union languages~\citep{steinberger2013dgt}. 

Data provided by the Department of Tourism, Culture, Arts, Gaeltacht, Sport and Media in Ireland formed the majority of the data in the public administration dataset. This includes staff notices, annual reports, website content, press releases and official correspondence. 

Parallel texts from the Digital Corpus of the European Parliament (DCEP) and the DGT are included in the training data. Crawled data, from sites of a similar domain are included. Furthermore a parallel corpus collected from Conradh na Gaeilge (CnaG), an Irish language organisation that promotes the Irish language, was included. The dataset was compiled as part of a previous study which carried out a preliminary comparison of SMT and NMT models for the Irish language ~\citep{dowling2018smt}. 

\subsubsection{Infrastructure}
Models were developed using a lab of machines each of which has an AMD Ryzen 7 2700X processor, 16 GB memory, a 256 SSD and an NVIDIA GeForce GTX 1080 Ti. Rapid prototype development was enabled through a Google Colab Pro subscription using NVIDIA Tesla P100 PCIe 16 GB graphic cards and up to 27GB of memory when available~\citep{bisong2019google}.

Our MT models were trained using the Pytorch implementation of OpenNMT 2.0, an open-source toolkit for NMT~\citep{klein2017opennmt}. 

\subsubsection{Metrics}

As part of this study, several automated metrics were used to determine the translation quality. All models were trained and evaluated on both the DGT and PA datasets using the BLEU~\citep{papineni2002BLEU}, TER~\citep{snover2006study} and ChrF~\citep{popovic2015ChrF} evaluation metrics. Case-insensitive BLEU scores, at the corpus level,  are reported. Model training was stopped once an early stopping criteria of no improvement in validation accuracy for 4 consecutive iterations was recorded.

\subsection{Results}

\subsubsection{Performance of subword models}
The impact on translation accuracy when choosing a subword model is highlighted in Tables \ref{tab:dgtvanilla-table} - \ref{tab:patrans-table}.  In training both RNN and Transformer architectures, incorporating any submodel type led to improvements in model accuracy. This finding is evident when training either the smaller generic DGT dataset or the larger in-domain PA dataset. 

Using an RNN architecture on DGT, as illustrated in Table \ref{tab:dgtvanilla-table}, the best performing model with a 32k unigram submodel, achieved a BLEU score 7.4\% higher than the baseline.  With the PA dataset using an RNN, as shown in Table 3, the model with the best BLEU, TER and ChrF3 scores again used a unigram submodel.

\begin{table}
\centering
\begin{tabular}{lllllllll}
\hline
\textbf{Architecture} &
  \textbf{BLEU} $\uparrow$ &
  \textbf{TER} $\downarrow$ &
  \textbf{ChrF3} $\uparrow$ &
  \textbf{Steps} &
  \textbf{\begin{tabular}[c]{@{}l@{}}Runtime \\ (hours)\end{tabular}} &
  \textbf{kgCO\textsubscript2} \\ \hline
dgt-rnn-base    & 52.7       & 0.42  & 0.71 & 75k  & 4.47 & 0 \\
dgt-rnn-bpe8k   & 54.6       & 0.40 & 0.73 & 85k  & 5.07 & 0 \\
dgt-rnn-bpe16k  & 55.6 & 0.39 & 0.74 & 100k & 5.58 & 0 \\
dgt-rnn-bpe32k  & 55.3       & 0.39 & 0.74 & 95k  & 4.67 & 0 \\ 
dgt-rnn-unigram & \textbf{55.6} & \textbf{0.39} & \textbf{0.74} & 105k & 5.07 & 0 \\ \hline
\end{tabular}
\caption{RNN performance on DGT dataset of 52k lines}
\label{tab:dgtvanilla-table}
\end{table}

\begin{table}
\centering
\begin{tabular}{lllllllll}
\hline
\textbf{Architecture} &
  \textbf{BLEU} $\uparrow$ &
  \textbf{TER} $\downarrow$ &
  \textbf{ChrF3} $\uparrow$ &
  \textbf{Steps} &
  \textbf{\begin{tabular}[c]{@{}l@{}}Runtime \\ (hours)\end{tabular}} &
  \textbf{kgCO\textsubscript2} \\ \hline
pa-rnn-base     & 40.4 & 0.47 & 0.63 & 60k  & 2.13 & 0 \\
pa-rnn-bpe8k   & 41.5 & 0.46 & 0.64 & 110k  & 4.16 & 0 \\
pa-rnn-bpe16k  & 41.5 & 0.46 & 0.64 & 105k & 3.78 & 0 \\
pa-rnn-bpe32k  & 41.9 & 0.47 & 0.64& 100k & 2.88 & 0 \\
pa-rnn-unigram & \textbf{41.9} & \textbf{0.46} & \textbf{0.64} & 95k & 2.75 & 0 \\ \hline
\end{tabular}
\caption{RNN performance on PA dataset of 88k lines}
\label{tab:pavanilla-table}
\end{table}

There are small improvements in BLEU scores when the RNN baseline is compared with  models using a BPE submodel of either 8k, 16k or 32k words, as illustrated in Tables \ref{tab:dgtvanilla-table} and \ref{tab:pavanilla-table}. The maximum BLEU score improvement of 1.5 points (2.5\%) is quite modest in the case of the public admin corpus.  However, there are larger gains with the DGT corpus. A baseline RNN model, trained on DGT,  achieved a BLEU score of 52.7 whereas the highest-performing BPE variant,  using a 16k vocab, recorded an improvement of nearly 3 points with a score of 55.6. 
\begin{table}
\centering
\begin{tabular}{lllllllll}
\hline
\textbf{Architecture} &
  \textbf{BLEU} $\uparrow$ &
  \textbf{TER} $\downarrow$ &
  \textbf{ChrF3} $\uparrow$ &
  \textbf{Steps} &
  \textbf{\begin{tabular}[c]{@{}l@{}}Runtime \\ (hours)\end{tabular}} &
  \textbf{kgCO\textsubscript2} \\ \hline
dgt-trans-base      & 53.4 & 0.41 & 0.72 & 55k  & 14.43 & 0.81 \\
dgt-trans-bpe8k     & 59.5 & 0.34 & 0.77 & 200k & 24.48 & 1.38 \\
dgt-trans-bpe16k    & \textbf{60.5} & \textbf{0.33} & \textbf{0.78} & 180k & 26.90 & 1.52 \\
dgt-trans-bpe32k    & 59.3 & 0.35 & 0.77 & 100k & 18.03 & 1.02 \\ 
dgt-trans-unigram   & 59.3 & 0.35  & 0.77 & 125k & 21.95 & 1.24 \\ \hline
\end{tabular}
\caption{Transformer performance on 52k DGT dataset. Highest performing model uses 2 attention heads. All other models use 8 attention heads.}
\label{tab:trans-table}
\end{table}

\begin{table}
\centering
\begin{tabular}{lllllllll}
\hline
\textbf{Architecture} &
  \textbf{BLEU} $\uparrow$ &
  \textbf{TER} $\downarrow$ &
  \textbf{ChrF3} $\uparrow$ &
  \textbf{Steps} &
  \textbf{\begin{tabular}[c]{@{}l@{}}Runtime \\ (hours)\end{tabular}} &
  \textbf{kgCO\textsubscript2} \\ \hline
pa-trans-base    & 44.1 & 0.44 & 0.66 & 20k  & 5.97  & 0.34 \\
pa-trans-bpe8k   & 46.6 & \textbf{0.40} & 0.68 & 160k & 20.1  & 1.13 \\
pa-trans-bpe16k  & \textbf{47.1} & 0.41 & \textbf{0.68} & 100k & 14.22 & 0.80  \\
pa-trans-bpe32k  & 46.8 & 0.41 & 0.68 & 70k  & 12.7  & 0.72 \\ 
pa-trans-unigram & 46.6 & 0.42 & 0.68 & 75k  & 13.34 & 0.75 \\ \hline
\end{tabular}
\caption{Transformer performance on 88k PA dataset. All models use 8 attention heads.}
\label{tab:patrans-table}
\end{table}

In the context of Transformer architectures, highlighted in Table \ref{tab:trans-table} and Table \ref{tab:patrans-table}, the use of subword models delivers significant performance improvements for both the DGT and public admin corpora. The performance gains for Transformer models are far greater than RNN models. Baseline DGT Transformer models achieve a BLEU score of 53.4 while a Transformer model, with a 16k BPE submodel, has a score of 60.5 representing a BLEU score improvement of 13\% at 7.1 BLEU points. 

For translating into a morphologically rich language, such as Irish, the ChrF metric has proven successful in showing strong correlation with human translation~\citep{stanojevic2015results}. In the context of our experiments, it worked well in highlighting the performance differences between RNN and Transformer architectures. 

\begin{figure}
    \centering
    \includegraphics[width=8cm]{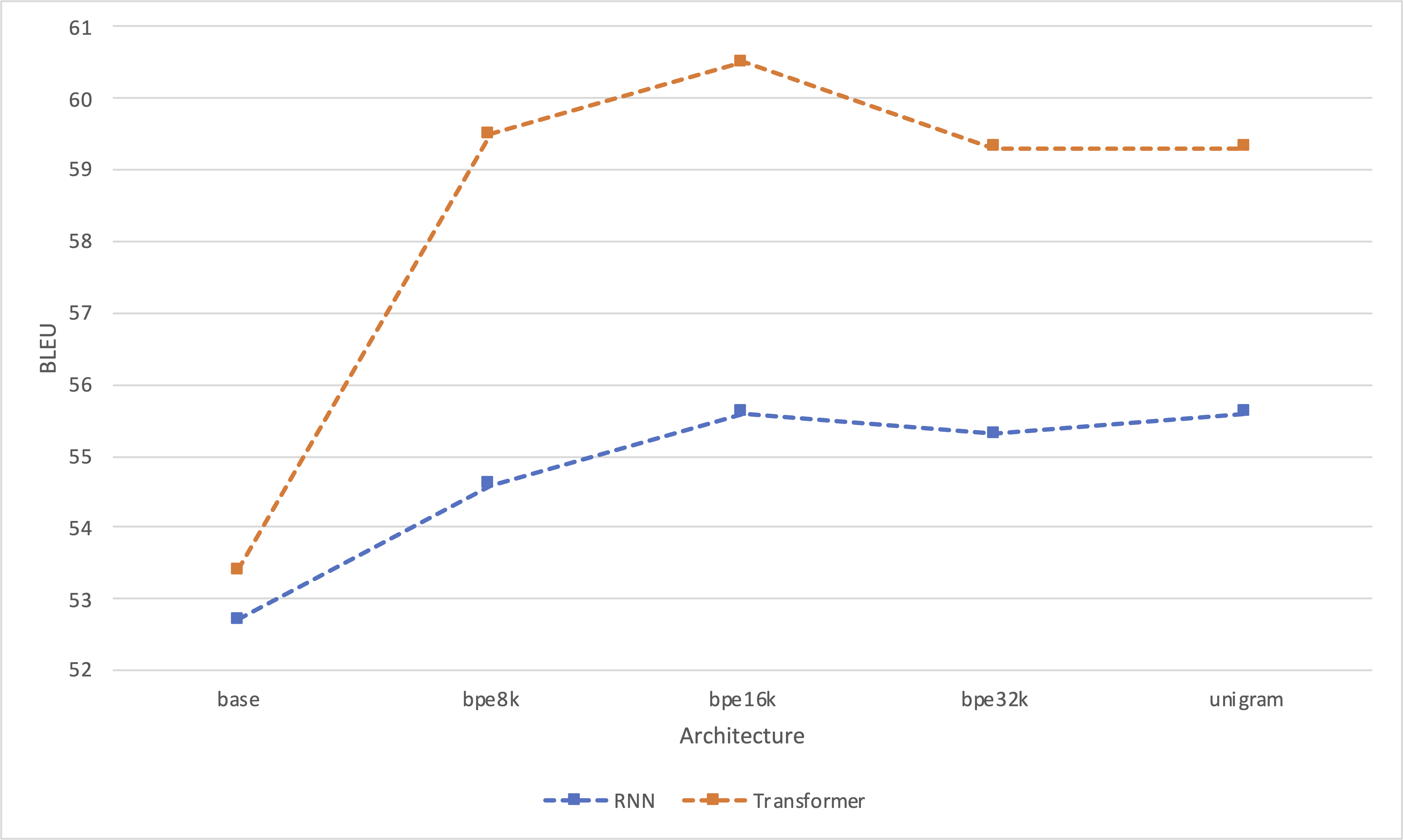}
    \caption{BLEU performance for all model architectures}
    \label{fig:dgt}
\end{figure}

\subsubsection{Transformer performance compared with RNN}

\begin{figure}[H]
    \centering
    \includegraphics[width=8cm]{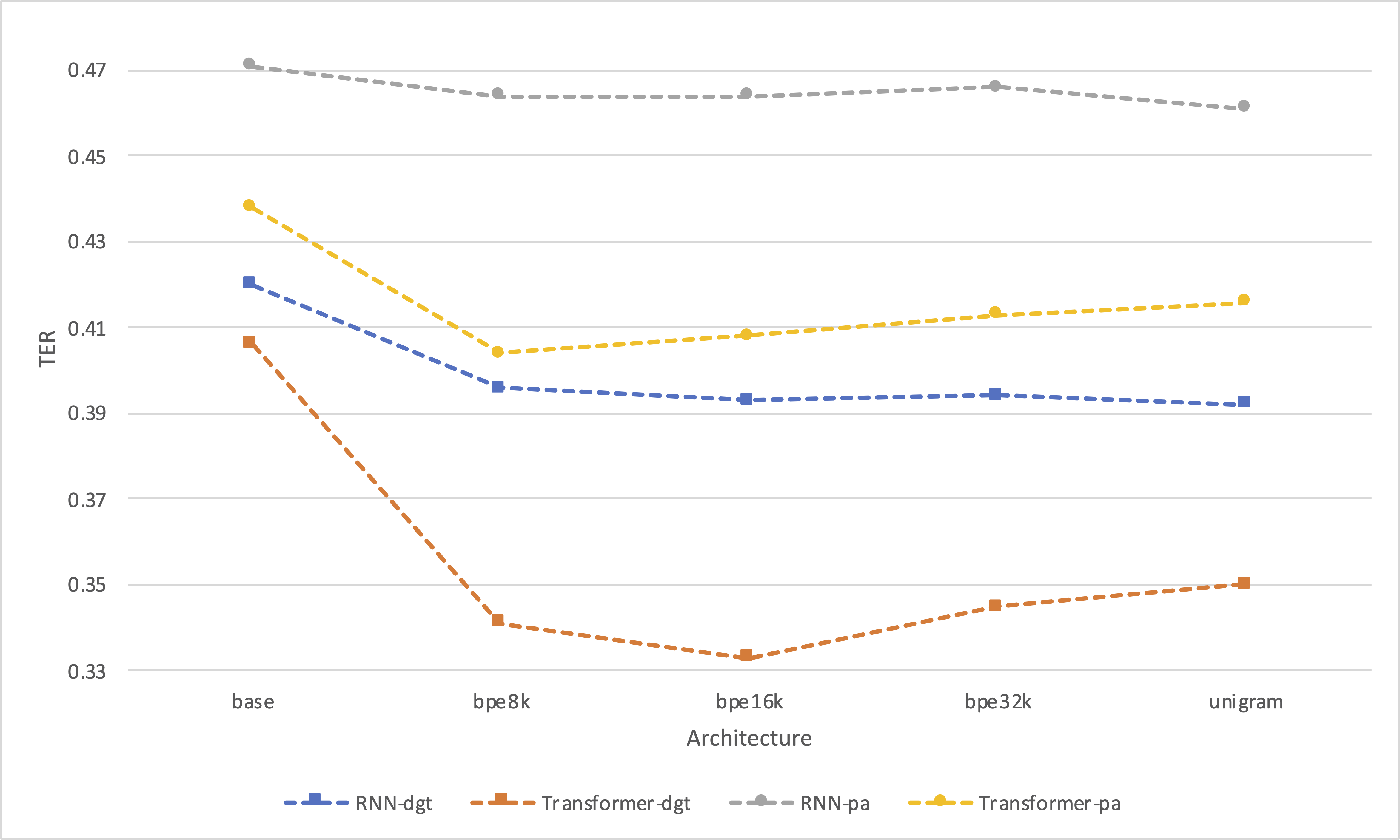}
    \caption{TER performance for all model architectures}
    \label{fig:pacompare}
\end{figure}

The performance of RNN models is contrasted with the Transformer approach in Figure \ref{fig:dgt} and Figure \ref{fig:pacompare}. Transformer models, as anticipated, outperform all their RNN counterparts. It is interesting to note the impact of choosing the optimal vocabulary size for BPE submodels. Both datasets demonstrate that choosing a BPE vocabulary of 16k words yields the highest performance. 

Furthermore, the TER scores highlighted in Figure \ref{fig:pacompare} reinforce the findings that using 16k BPE submodels on Transformer architectures leads to better translation performance. The TER score for the DGT Transformer 16k BPE model is significantly better (0.33) when compared with the baseline performance (0.41).

\begin{multicols}{2}

\begin{figure}[H]
\centering
    \includegraphics[width=4.5cm]{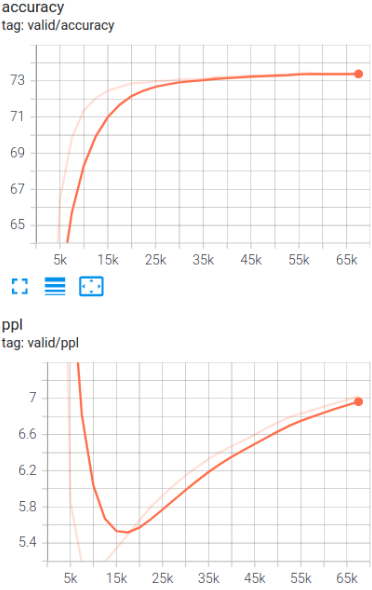}
    \caption{Training DGT Transformer baseline}
    \label{fig:train-base}
\end{figure}

\begin{figure}[H]
\centering
    \includegraphics[width=4.55cm]{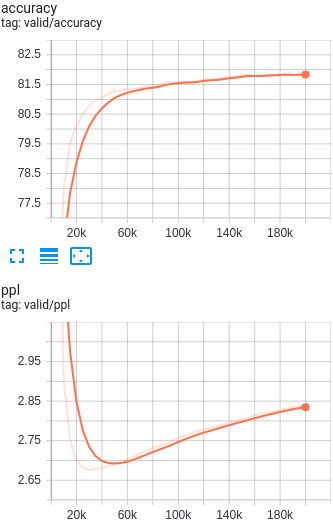}
    \caption{Training DGT Transformer 16k BPE}
    \label{fig:train-bpe}
\end{figure}

\end{multicols}

\section{Environmental Impact}
Motivated by the findings of Stochastic Parrots \citep{bender2021dangers}, energy consumption during model development was tracked. Prototype model development used Colab Pro, which as part of Google Cloud is carbon neutral \citep{lacoste2019quantifying}. However, longer running Transformer experiments were conducted on local servers using 324 gCO\textsubscript2 per kWh \footnote{https://www.seai.ie/publications/Energy-in-Ireland-2020.pdf}\citep{sei2020}. The net result was just under 10 kgCO\textsubscript2 created for a full run of model development. Models developed during this study, will be reused for ensemble experiments in future work.

\section{Discussion}
Validation accuracy, and model perplexity, in developing the baseline and optimal models for the DGT corpus are illustrated in Figure \ref{fig:train-base} and Figure \ref{fig:train-bpe}. Rapid convergence was observed while training the baseline model such that little accuracy improvement occurs after 20k steps. Including a subword model led to much slower convergence and there were only marginal gains after 60k steps. Furthermore, it is observed that training the DGT model, with a 16k BPE submodel, boosted validation accuracy by over 8\% compared with its baseline.

With regard to the key metric of perplexity, it is shown to rise after training for 15k steps in the baseline models. PPL was observed to rise at later stages, typically after 40k steps in models developed using subword models. Perplexity (PPL), shows how many different, equally probable words can be produced during translation. As a metric for translation performance, it is important to keep low scores  so the number of alternative translations is reduced. Therefore, for future model development it may be worthwhile to set PPL as an early stopping parameter.

On examining the PPL graphs of Figure \ref{fig:train-base} and Figure \ref{fig:train-bpe}, it is clear that a lower global minimum is achieved when the Transformer approach is used with a 16k BPE submodel. The PPL global minimum (2.7) is over 50\% lower than the corresponding PPL for the Transformer base model (5.5). Such a finding illustrates that choosing an optimal submodel delivers significant performance gains.

Translation engine performance was bench-marked against Google Translate's \footnote{https://translate.google.com/} English to Irish translation service which is freely available on the internet. Four random samples were selected from the English source test file and are presented in Table~\ref{tab:translations}. Translation of these samples was carried out on the optimal DGT Transformer model and using Google Translate. Case insensitive, sentence level BLEU scores were recorded and are presented in Table~\ref{tab:trans-google}. The results are encouraging and indicate well-performing translation models on the DGT dataset.

The optimal parameters selected in this discovery process are identified in bold in Table 2. A higher initial learning rate of 2 coupled with an average decay of 0.0001 led to longer training times but more accurate models. Despite setting an early stopping parameter, many of the Transformer builds continued for the full cycle of 200k steps over periods of 20+ hours. 

Training transformer models with a reduced number of attention heads led to a marginal improvement in translation accuracy with a smaller corpus. Our best performing model on a 55k DGT corpus, with 2 heads and a 16k BPE submodel, achieved a BLEU score of 60.5 and a TER score of 0.33. By comparison, using 8 heads with the same architecture and dataset yielded 60.3 for the BLEU and 0.34 for the TER. In the case of a larger 88k PA corpus, all transformer models using 8 heads performed better than equivalent models using just 2 heads.
\begin{table}[H]
\begin{tabular}{ll}
\textbf{Source Language (English)} & \textbf{Reference Human Translation (Irish)} \\  \hline
\begin{tabular}[c]{@{}l@{}}A clear harmonised procedure, including the \\ necessary criteria for disease–free status, \\ should be established for that purpose.\end{tabular} & \begin{tabular}[c]{@{}l@{}}Ba cheart nós imeachta comhchuibhithe soiléir, \\ lena n-áirítear na critéir is gá do stádas saor \\ ó ghalar, a bhunú chun na críche sin.\end{tabular} \\ \\
\begin{tabular}[c]{@{}l@{}}the mark is applied anew, as appropriate.\end{tabular} & \begin{tabular}[c]{@{}l@{}}déanfar an mharcáil arís, mar is iomchuí.\end{tabular} \\ \\
\begin{tabular}[c]{@{}l@{}}If the court decides that a review is \\ justified on any of the grounds set out in \\ paragraph 1, the judgment given in the \\ European Small Claims Procedure shall \\ be null and void.\end{tabular} & \begin{tabular}[c]{@{}l@{}}Má chinneann an chúirt go bhfuil bonn cirt \\ le hathbhreithniú de bharr aon cheann de na \\ forais a leagtar amach i mír 1, beidh an \\ breithiúnas a tugadh sa Nós Imeachta Eorpach \\ um Éilimh Bheaga ar neamhní go hiomlán.\end{tabular} \\ \\
households where pet animals are kept; & teaghlaigh ina gcoimeádtar peataí; \\ \hline

\end{tabular}
\caption{Samples of human reference translations}
\label{tab:translations}
\end{table}

\begin{table}[H]
\begin{tabular}{lclc}
\textbf{Transformer (16 kBPE)} & \textbf{BLEU} $\uparrow$ & \textbf{Google Translate} & \textbf{BLEU} $\uparrow$\\ \hline

\begin{tabular}[c]{@{}l@{}}Ba cheart nós imeachta soiléir \\ comhchuibhithe, lena n-áirítear \\ na critéir is gá maidir le \\ stádas saor ó ghalair, a bhunú \\ chun na críche sin.
\end{tabular} & 61.6 & \begin{tabular}[c]{@{}l@{}}Ba cheart nós imeachta \\ comhchuibhithe soiléir, lena \\ n-áirítear na critéir riachtanacha \\ maidir le stádas saor ó ghalair, \\ a bhunú chun na críche sin.\end{tabular} & 70.2 \\ \\ 
\begin{tabular}[c]{@{}l@{}}go gcuirtear an marc i bhfeidhme,\\ de réir mar is iomchuí.\end{tabular} & 21.4 & \begin{tabular}[c]{@{}l@{}}cuirtear an marc i bhfeidhm as \\ an nua, de réir mar is cuí.\end{tabular} & 6.6 \\ \\
\begin{tabular}[c]{@{}l@{}}
Má chinneann an chúirt go bhfuil \\ bonn cirt le hathbhreithniú ar aon \\ cheann de na forais a leagtar amach \\ i mír 1, beidh an breithiúnas a \\ thugtar sa Nós Imeachta Eorpach \\ um Éilimh Bheaga ar neamhní.

\end{tabular} & 77.3 & \begin{tabular}[c]{@{}l@{}}Má chinneann an chúirt go bhfuil \\ údar le hathbhreithniú ar aon \\ cheann de na forais atá leagtha \\ amach i mír 1, beidh an \\ breithiúnas a thugtar  sa \\ Nós Imeachta Eorpach um \\ Éilimh Bheaga ar neamhní\end{tabular} & 59.1 \\ \\
teaghlaigh ina gcoimeádtar peataí; & 100 & teaghlaigh ina gcoinnítear peataí; & 30.2 \\ \hline
\end{tabular}
\caption{Transformer model compared with Google Translate using random samples from the DGT corpus. Full evaluation of Google Translate on the DGT test set, with 1.3k lines, generated a BLEU score of 46.3 and a TER score of 0.44. Comparative scores on the test set using our Transformer model, with 2 attention heads and 16k BPE submodel realised 60.5 for BLEU and 0.33 for TER.}
\label{tab:trans-google}
\end{table}

Standard Transformer parameters for batch size (2048) and the number of encoder / decoder layers (6) were all observed to perform well on the DGT and PA corpora. Reducing hidden neurons to 256 and increasing regularization dropout to 0.3 improved translation performance and were chosen when building all Transformer models.

\section{Conclusion}

In our paper, we demonstrated that a random search approach to hyperparameter optimization leads to the development of high-performing translation models. 

We have shown that choosing subword models, in our low-resource scenarios, is an important driver for the performance of MT engines. Moreover, the choice of vocabulary size leads to varying degrees of performance. Within the context of low-resource English to Irish translation, we achieved optimal performance, on a 55k generic corpus and an 88k in-domain corpus, when a Transformer architecture with a 16k BPE submodel was used. 
The importance of selecting hyperparameters in training low-resource Transformer models was also demonstrated. By reducing the number of hidden layer neurons and increasing dropout, our models performed significantly better than baseline models and Google Translate.

Performance improvement of our optimized Transformer models, with subword segmentation, was observed across all key indicators namely a higher validation accuracy, a PPL achieved at a lower global minimum, a lower post editing effort and a higher translation accuracy.  

\section*{Acknowledgements}
This work was supported by ADAPT,   which is funded under the SFI Research Centres Programme (Grant 13/RC/2016) and is co-funded by the European Regional Development Fund. This research was also funded by the Munster Technological University.
\small

\bibliographystyle{apalike}
\bibliography{mtsummit2021}

\begin{thebibliography}{}

\bibitem[Alam and Imran, 2015]{alam2015digital}
Alam, K. and Imran, S. (2015).
\newblock The digital divide and social inclusion among refugee migrants.
\newblock {\em Information Technology \& People}.

\bibitem[Araabi and Monz, 2020]{araabi2020optimizing}
Araabi, A. and Monz, C. (2020).
\newblock Optimizing transformer for low-resource neural machine translation.
\newblock {\em arXiv preprint arXiv:2011.02266}.

\bibitem[Bahdanau et~al., 2014]{bahdanau2014neural}
Bahdanau, D., Cho, K., and Bengio, Y. (2014).
\newblock Neural machine translation by jointly learning to align and translate.
\newblock {\em arXiv preprint arXiv:1409.0473}.

\bibitem[Bender et~al., 2021]{bender2021dangers}
Bender, E.~M., Gebru, T., McMillan-Major, A., and Shmitchell, S. (2021).
\newblock On the dangers of stochastic parrots: Can language models be too big?
\newblock In {\em Proceedings of the 2021 ACM Conference on Fairness, Accountability, and Transparency}, pages 610--623.

\bibitem[Bergstra and Bengio, 2012]{bergstra2012random}
Bergstra, J. and Bengio, Y. (2012).
\newblock Random search for hyper-parameter optimization.
\newblock {\em Journal of machine learning research}, 13(2).

\bibitem[Bisong, 2019]{bisong2019google}
Bisong, E. (2019).
\newblock Google colaboratory.
\newblock In {\em Building Machine Learning and Deep Learning Models on Google Cloud Platform}, pages 59--64. Springer.

\bibitem[Bojar et~al., 2017]{bojar-etal-2017-findings}
Bojar, O., Chatterjee, R., Federmann, C., Graham, Y., Haddow, B., Huang, S., Huck, M., Koehn, P., Liu, Q., Logacheva, V., Monz, C., Negri, M., Post, M., Rubino, R., Specia, L., and Turchi, M. (2017).
\newblock Findings of the 2017 conference on machine translation ({WMT}17).
\newblock In {\em Proceedings of the Second Conference on Machine Translation}, pages 169--214, Copenhagen, Denmark. Association for Computational Linguistics.

\bibitem[Bojar et~al., 2018]{bojar-etal-2018-findings}
Bojar, O., Federmann, C., Fishel, M., Graham, Y., Haddow, B., Koehn, P., and Monz, C. (2018).
\newblock Findings of the 2018 conference on machine translation ({WMT}18).
\newblock In {\em Proceedings of the Third Conference on Machine Translation: Shared Task Papers}, pages 272--303, Belgium, Brussels. Association for Computational Linguistics.

\bibitem[Cho et~al., 2014]{cho2014properties}
Cho, K., Van~Merri{\"e}nboer, B., Bahdanau, D., and Bengio, Y. (2014).
\newblock On the properties of neural machine translation: Encoder-decoder approaches.
\newblock {\em arXiv preprint arXiv:1409.1259}.

\bibitem[Ding et~al., 2019]{ding2019call}
Ding, S., Renduchintala, A., and Duh, K. (2019).
\newblock A call for prudent choice of subword merge operations in neural machine translation.
\newblock {\em arXiv preprint arXiv:1905.10453}.

\bibitem[Dowling et~al., 2018]{dowling2018smt}
Dowling, M., Lynn, T., Poncelas, A., and Way, A. (2018).
\newblock Smt versus nmt: Preliminary comparisons for irish.

\bibitem[Gage, 1994]{gage1994new}
Gage, P. (1994).
\newblock A new algorithm for data compression.
\newblock {\em C Users Journal}, 12(2):23--38.

\bibitem[Ghifary et~al., 2016]{ghifary2016deep}
Ghifary, M., Kleijn, W.~B., Zhang, M., Balduzzi, D., and Li, W. (2016).
\newblock Deep reconstruction-classification networks for unsupervised domain adaptation.
\newblock In {\em European Conference on Computer Vision}, pages 597--613. Springer.

\bibitem[Gowda and May, 2020]{gowda2020finding}
Gowda, T. and May, J. (2020).
\newblock Finding the optimal vocabulary size for neural machine translation.
\newblock {\em arXiv preprint arXiv:2004.02334}.

\bibitem[Klein et~al., 2017]{klein2017opennmt}
Klein, G., Kim, Y., Deng, Y., Senellart, J., and Rush, A.~M. (2017).
\newblock Opennmt: Open-source toolkit for neural machine translation.
\newblock {\em arXiv preprint arXiv:1701.02810}.

\bibitem[Kudo, 2018]{kudo2018subword}
Kudo, T. (2018).
\newblock Subword regularization: Improving neural network translation models with multiple subword candidates.
\newblock {\em arXiv preprint arXiv:1804.10959}.

\bibitem[Kudo and Richardson, 2018]{kudo2018SentencePiece}
Kudo, T. and Richardson, J. (2018).
\newblock Sentencepiece: A simple and language independent subword tokenizer and detokenizer for neural text processing.
\newblock {\em arXiv preprint arXiv:1808.06226}.

\bibitem[Lacoste et~al., 2019]{lacoste2019quantifying}
Lacoste, A., Luccioni, A., Schmidt, V., and Dandres, T. (2019).
\newblock Quantifying the carbon emissions of machine learning.
\newblock {\em arXiv preprint arXiv:1910.09700}.

\bibitem[Liu et~al., 2018]{liu2018pivot}
Liu, C.-H., Silva, C.~C., Wang, L., and Way, A. (2018).
\newblock Pivot machine translation using chinese as pivot language.
\newblock In {\em China Workshop on Machine Translation}, pages 74--85. Springer.

\bibitem[MacFarlane et~al., 2008]{macfarlane2008responses}
MacFarlane, A., Glynn, L.~G., Mosinkie, P.~I., and Murphy, A.~W. (2008).
\newblock Responses to language barriers in consultations with refugees and asylum seekers: a telephone survey of irish general practitioners.
\newblock {\em BMC Family Practice}, 9(1):1--6.

\bibitem[Montgomery, 2017]{montgomery2017design}
Montgomery, D.~C. (2017).
\newblock {\em Design and analysis of experiments}.
\newblock John wiley \& sons.

\bibitem[Papineni et~al., 2002]{papineni2002BLEU}
Papineni, K., Roukos, S., Ward, T., and Zhu, W.-J. (2002).
\newblock Bleu: a method for automatic evaluation of machine translation.
\newblock In {\em Proceedings of the 40th annual meeting of the Association for Computational Linguistics}, pages 311--318.

\bibitem[Popovi{\'c}, 2015]{popovic2015ChrF}
Popovi{\'c}, M. (2015).
\newblock chrf: character n-gram f-score for automatic mt evaluation.
\newblock In {\em Proceedings of the Tenth Workshop on Statistical Machine Translation}, pages 392--395.

\bibitem[Sanders and Giraud-Carrier, 2017]{sanders2017informing}
Sanders, S. and Giraud-Carrier, C. (2017).
\newblock Informing the use of hyperparameter optimization through metalearning.
\newblock In {\em 2017 IEEE International Conference on Data Mining (ICDM)}, pages 1051--1056. IEEE.

\bibitem[SEAI, 2020]{sei2020}
SEAI (2020).
\newblock {Sustainable Energy in Ireland}.

\bibitem[Sennrich et~al., 2015]{sennrich2015neural}
Sennrich, R., Haddow, B., and Birch, A. (2015).
\newblock Neural machine translation of rare words with subword units.
\newblock {\em arXiv preprint arXiv:1508.07909}.

\bibitem[Snover et~al., 2006]{snover2006study}
Snover, M., Dorr, B., Schwartz, R., Micciulla, L., and Makhoul, J. (2006).
\newblock A study of translation edit rate with targeted human annotation.
\newblock In {\em Proceedings of association for machine translation in the Americas}, volume 200. Citeseer.

\bibitem[Stanojevi{\'c} et~al., 2015]{stanojevic2015results}
Stanojevi{\'c}, M., Kamran, A., Koehn, P., and Bojar, O. (2015).
\newblock Results of the wmt15 metrics shared task.
\newblock In {\em Proceedings of the Tenth Workshop on Statistical Machine Translation}, pages 256--273.

\bibitem[Steinberger et~al., 2013]{steinberger2013dgt}
Steinberger, R., Eisele, A., Klocek, S., Pilos, S., and Schl{\"u}ter, P. (2013).
\newblock Dgt-tm: A freely available translation memory in 22 languages.
\newblock {\em arXiv preprint arXiv:1309.5226}.

\bibitem[Van~Biljon et~al., 2020]{van2020optimal}
Van~Biljon, E., Pretorius, A., and Kreutzer, J. (2020).
\newblock On optimal transformer depth for low-resource language translation.
\newblock {\em arXiv preprint arXiv:2004.04418}.

\bibitem[Vaswani et~al., 2017]{vaswani2017attention}
Vaswani, A., Shazeer, N., Parmar, N., Uszkoreit, J., Jones, L., Gomez, A.~N., Kaiser, L., and Polosukhin, I. (2017).
\newblock Attention is all you need.
\newblock {\em arXiv preprint arXiv:1706.03762}.

\bibitem[Way, 2020]{way_2020}
Way, A. (2020).
\newblock {MT Developments in the EU: Keynote AMTA 2020}.

\end{thebibliography}

\end{document}